\documentclass[a4paper, 10pt, conference]{ieeeconf}      % Use this line for a4 paper

\IEEEoverridecommandlockouts                              % This command is only needed if 
\usepackage{graphics} % for pdf, bitmapped graphics files
\usepackage{epsfig} % for postscript graphics files
\usepackage{amsmath} % assumes amsmath package installed
\usepackage{amssymb}  % assumes amsmath package installed
\usepackage{lipsum}
\usepackage{algorithm}
\usepackage{algpseudocode}
\usepackage{esvect}
\usepackage{cite}
\usepackage{subfig}
\usepackage{url}
\usepackage[flushleft]{threeparttable}

\algnewcommand\algorithmicforeach{\textbf{for each}}
\algdef{S}[FOR]{ForEach}[1]{\algorithmicforeach\ #1\ \algorithmicdo}

\usepackage{tikz}

\title{\LARGE \bf
A Realistic Surgical Simulator for Non-Rigid and Contact-Rich Manipulation in Surgeries with the da Vinci Research Kit
}

\author{Yafei Ou, \textit{Student Member, IEEE}, Sadra Zargarzadeh$^{*}$, \textit{Student Member, IEEE},\\Paniz Sedighi$^{*}$, \textit{Student Member, IEEE}, and Mahdi Tavakoli, \textit{Senior Member, IEEE}% <-this % stops a space
\thanks{This research was supported by the Canada Foundation for Innovation (CFI), the Natural Sciences and Engineering Research Council (NSERC) of Canada, the Canadian Institutes of Health Research (CIHR), Alberta Innovates, the Alberta Jobs, Economy, and Trade Ministry’s Major Initiatives Fund A-Medico, China Scholarship Council (CSC), and Alberta Advanced Education. \textit{(Corresponding author: Yafei Ou)}}% <-this % stops a space
\thanks{Yafei Ou, Sadra Zargarzadeh, and Paniz Sedighi are with the Department of Electrical and Computer Engineering, University of Alberta, Edmonton, Alberta, Canada. {\tt\footnotesize \{yafei.ou, sadra.zar, sedighi1\}@ualberta.ca}}%
\thanks{Mahdi Tavakoli is with the Department of Electrical and Computer Engineering and the Department of Biomedical Engineering, University of Alberta, Edmonton, Alberta, Canada. {\tt\footnotesize mahdi.tavakoli@ualberta.ca}}%
\thanks{*Sadra Zargarzadeh and Paniz Sedighi contributed equally.}
}

\def\BibTeX{{\rm B\kern-.05em{\sc i\kern-.025em b}\kern-.08em
    T\kern-.1667em\lower.7ex\hbox{E}\kern-.125emX}}

\begin{document}

\maketitle
\thispagestyle{empty}
\pagestyle{empty}

% Floating Text Box
\begin{tikzpicture}[remember picture, overlay]
\node [align=left, xshift=10.5cm, yshift=-1.5cm] at (current page.north west) % Positioning
{
\begin{minipage}{19cm} % Adjust the width of the text box
% \fbox{% You can remove \fbox if you don't want a border
\footnotesize
Accepted for presentation at the 21st International Conference on Ubiquitous Robots (UR), New York, USA
\\ \\
\copyright 2024 IEEE. Personal use of this material is permitted. Permission from IEEE must be obtained for all other uses, in any current or future media, including reprinting/republishing this material for advertising or promotional purposes, creating new collective works, for resale or redistribution to servers or lists, or reuse of any copyrighted component of this work in other works.
\\ \\
DOI: 10.1109/UR61395.2024.10597513
% }
\end{minipage}
};
\end{tikzpicture}

%%%%%%%%%%%%%%%%%%%%%%%%%%%%%%%%%%%%%%%%%%%%%%%%%%%%%%%%%%%%%%%%%%%%%%%%%%%%%%%%
\begin{abstract}

Realistic real-time surgical simulators play an increasingly important role in surgical robotics research, such as surgical robot learning and automation, and surgical skills assessment. Although there are a number of existing surgical simulators for research, they generally lack the ability to simulate the diverse types of objects and contact-rich manipulation tasks typically present in surgeries, such as tissue cutting and blood suction. In this work, we introduce CRESSim, a realistic surgical simulator based on PhysX 5 for the da Vinci Research Kit (dVRK) that enables simulating various contact-rich surgical tasks involving different surgical instruments, soft tissue, and body fluids. The real-world dVRK console and the master tool manipulator (MTM) robots are incorporated into the system to allow for teleoperation through virtual reality (VR). To showcase the advantages and potentials of the simulator, we present three examples of surgical tasks, including tissue grasping and deformation, blood suction, and tissue cutting. These tasks are performed using the simulated surgical instruments, including the large needle driver, suction irrigator, and curved scissor, through VR-based teleoperation.

\end{abstract}

%%%%%%%%%%%%%%%%%%%%%%%%%%%%%%%%%%%%%%%%%%%%%%%%%%%%%%%%%%%%%%%%%%%%%%%%%%%%%%%%
\section{Introduction}

High-performance and realistic real-time simulators are playing an increasingly significant role in robotics research, not only because they serve as a playground for testing control and automation algorithms without the need for real robots, but also because a large number of recent advances using machine learning (ML) approaches are heavily driven by simulators that provide synthetic data.

However, developing real-time simulators for surgical robots poses unique challenges compared with other general robotics scenarios, due to the nature of surgeries that require a diverse and unique range of contact-rich manipulation. Unlike general robotics such as autonomous driving and home service robots, multiple types of objects, including rigid bodies (e.g. surgical instruments), soft bodies (e.g. voluminous soft tissue), fluids (e.g. blood and other body fluids), and cloth-type tissue (e.g. fascia) are commonly present in one single surgical scene, contacting with each other. Furthermore, complex manipulations, such as cutting, cauterization, and fluid suctioning do not usually exist in other general robotics simulation scenes. These factors lead to the need for specific adaptations and designs for surgical simulators. For instance, to simulate cauterization, burning and smoking effects must be included.

\begin{figure}[t]
    \centering
    \includegraphics[width=0.8\columnwidth]{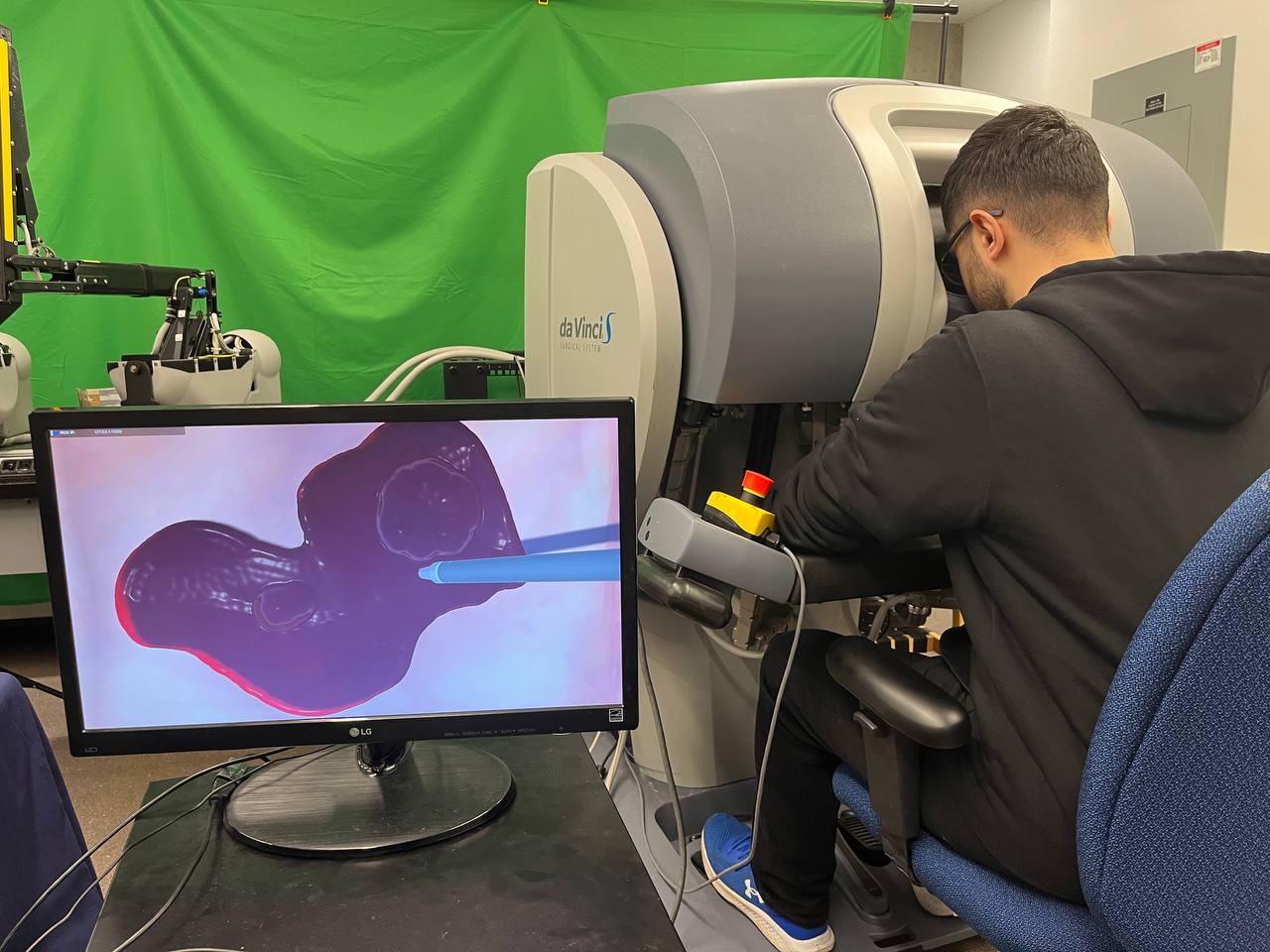}
    \caption{Teleoperating the CRESSim environment using the dVRK console.}
    \label{fig:teleoperation_demo}%
\end{figure}

With the help of the da Vinci Research Kit (dVRK) and its open-sourced software \cite{kazanzides2014open}, surgical robotics research has gained increasing attention. While there are a number of existing surgical simulators featuring the da Vinci system, they are either only for commercial surgical training purposes or limited to specific scenarios and cannot simulate various types of objects presented typically in a surgery. The da Vinci SimNow\footnote{\url{https://www.intuitive.com/products-and-services/da-vinci/learning/simnow/}} is a commercial surgical skills training simulator developed by Intuitive Surgical, Inc.~(Sunnyvale, California). Although it supports a variety of training tasks such as vessel clipping and tissue cutting, it is only for training surgeons using proprietary robotic platforms and does not provide an open interface for robotics research. Asynchronous Multi-Body Framework (AMBF) \cite{munawar2019real} is an open-sourced simulator based on Bullet Physics \cite{coumans2016pybullet}, which supports rigid and soft body simulation through finite element method (FEM). However, while Bullet provides high-performance rigid-body simulation, the FEM soft body is less realistic. Furthermore, it is unable to simulate fluids, which usually exist in surgical environments such as blood and other body fluids. Similarly, Assisted Teleoperation with Augmented Reality (ATAR) \cite{enayati2018robotic} and SurRoL \cite{xu2021surrol} are also based on Bullet, which have the same limitations.

Some other surgical simulators are developed based on the Simulation Open Framework Architecture\footnote{\url{https://www.sofa-framework.org/}} (SOFA), such as LapGym \cite{scheikl2023lapgym}. However, SOFA sacrifices computation costs for a more accurate FEM performance than Bullet, making simulating large-scale and complex surgical scenes slow. Furthermore, it does not support fluids as well. UnityFlexML, on the other hand, utilizes Nvidia FleX for Unity, a position-based dynamics (PBD) engine plugin for Unity, to simulate soft tissue manipulation \cite{tagliabue2020soft}. While diverse objects including rigid bodies, soft bodies, cloths, and fluids are natively supported by the FleX engine, the discontinuation of FleX for Unity makes it inadvisable and inconvenient to develop new environments and new manipulations such as cutting. Furthermore, PBD soft bodies are less realistic compared with FEM soft bodies, especially when large position displacements or velocities are applied to the particles.

In this work, we present \textbf{CRESSim}, a \textbf{C}ontact-\textbf{R}ich \textbf{E}nvironment for \textbf{S}urgical \textbf{Sim}ulation. CRESSim is a realistic simulator that enables the simulation of various contact-rich manipulation tasks in surgeries. The developed system is built on Unity and PhysX 5 SDK, allowing the simulation of rigid bodies, serial robots, soft bodies, cloth, fluid, and complex manipulation tasks such as cutting. The main contributions are as follows:
\begin{itemize}
    \item We build a new platform for surgical simulation featuring the dVRK using Unity and PhysX 5 SDK. A Unity native plugin is developed to enable GUI-based interactive editing of simulation scenes, allowing adding and removing various objects and constraints.
    \item We introduce two new simulated surgical instruments and three contact-rich surgical tasks in the proposed simulator. These include tissue grasping and manipulation, blood suction (single-arm tasks), and tissue cutting (a bimanual task).
    \item We incorporate the real-world dVRK console to allow teleoperating the simulated robots through virtual reality (VR), and present preliminary demonstrations for completing the three contact-rich tasks using teleoperation.
\end{itemize}
To the best of the authors' knowledge, this is also the first time that the PhysX 5 library has been integrated into Unity for building a surgical simulation platform.

\newcommand\Tstrut{\rule{0pt}{2.6ex}}         % = `top' strut
\newcommand\Bstrut{\rule[-0.9ex]{0pt}{0pt}}   % = `bottom' strut

\begin{table*}[t]
\caption{Comparison of CRESSim with existing surgical simulators}
\label{tab:comparison}
\centering
\begin{threeparttable}
    \begin{tabular}{cccc}
    \hline
    Platform                                      & Physics$^1$                              & Manipulation$^2$                 & Robot integration$^3$  \Tstrut \Bstrut\\ \hline
    \Tstrut AMBF, ATAR (Bullet)                           & R, S$_F$                             & R, S                         & D$^+$, L, T          \\
    SurRoL, V-Rep Simulator for the dVRK (Bullet) & R, S$_F^*$                           & R, S$^*$                     & D$^+$, L, T          \\
    LapGym (SOFA)                                 & R, S$_F$, C$_F$                             & R, S, C, C$^+$               & G                 \\
    UnityFlexML (FleX)                            & R$_P$, S$_P$, C$_P$, F$_P$           & R, S, C$^*$, C$^{+*}$  F$^*$ & D, L, T              \Bstrut\\ \hline
    CRESSim (PhysX 5)                            & R, R$_P^*$, S$_F$, S$_P^*$, C$_F^*$, C$_P$, F$_P$ & R, S, C, C$^+$, F            & D$^+$, L, C, S, T         \Tstrut\Bstrut\\ \hline
    \end{tabular}
    \begin{tablenotes}
      \small
      \item[1] Physics: regular rigid body (R), PBD rigid body (R$_P$), FEM soft body (S$_F$), PBD soft body (S$_P$), FEM cloth (C$_F$), PBD cloth (C$_P$), and position-based fluids (F$_P$).
      \item[2] Manipulation: rigid object grasping (R), soft object grasping and deformation (S), cloth grasping and deformation (C), cloth cutting (C$^+$), and fluid manipulation (F).
      \item[3] Robot integration: kinematic-only dVRK robots (D), dVRK robots with dynamics (D$^+$), large needle driver (L), curved scissor (C), suction irrigator (S), general laparoscopic tools (G), and teleoperation (T).
      \item[$*$] Items marked with $*$: it is technically possible but has not been implemented or used in the platform.
    \end{tablenotes}
\end{threeparttable}
\end{table*}

\section{Related Work}

\subsection{Surgery and Surgical Robot Simulation}
Existing surgical simulators can be categorized into commercial ones such as the da Vinci SimNow and VirtaMed LaparoS\footnote{\url{https://www.virtamed.com/products-and-solutions/simulators/laparos}} which are mainly used for training surgeons, and open-source research platforms such as LapGym \cite{scheikl2023lapgym}, AMBF \cite{munawar2019real}, ATAR \cite{enayati2018robotic}, and V-Rep Simulator for the dVRK \cite{fontanelli2018v}, that focus on providing a multi-faceted interface for robotics research. Other surgical simulators such as SurRoL \cite{xu2021surrol}, dVRL \cite{richter2019open}, AMBF-RL \cite{varier2022ambf}, and UnityFlexML \cite{tagliabue2020soft} have specific focuses on machine learning applications. Table~\ref{tab:comparison} shows a comparison between the proposed simulation environment and the existing surgical simulators.

\subsection{Simulation for Non-rigid and Contact-Rich Manipulation}
There are a number of simulators for non-rigid and contact-rich manipulation in the general robotics field. However, most of them are specific to a particular type of manipulation, such as fluid manipulation. To simulate fluids, \cite{gu2023maniskill2} incorporates the material point method (MPM) into SAPIEN \cite{xiang2020sapien}, a PhysX 4-based simulator that only supports rigid body dynamics. FluidLab \cite{xian2023fluidlab} proposes FluidEngine based on MPM that covers a wider range of fluids. SoftGym \cite{lin2021softgym} builds on top of Nvidia FleX, and TDW \cite{gan2020threedworld} integrates PhysX 4 with Nvidia FleX to support soft bodies, cloth, and fluids. However, due to the discontinuation of FleX and Flex for Unity, it is inconvenient to develop new simulators based on it, especially when developing customized manipulation behavior such as cutting. DiffSim \cite{qiao2020scalable} specializes in soft-body deformation, such as cloth folding. DiSECt \cite{heiden2021disect} has a specific focus on cutting voluminous soft materials using FEM.

The advancement of simulators has contributed to the development of methods for the automatic control of robots in completing contact-rich daily tasks, such as pouring \cite{babaians2022pournet}, cloth folding \cite{weng2022fabricflownet,ha2022flingbot}, and soft object deformation \cite{ficuciello2018fem}. Although these studies represent progress in simulation environments and robot automation in the general robotics field, they are specific to a limited range of objects and manipulation types. However, surgical scenes require the presence of various types of objects and the simulation of different manipulation tasks (grasping, cutting, fluid suction, burning, and more) in a single scene, making it impractical to directly use these simulators for surgical simulation.

\begin{figure}[t]
    \centering
    \includegraphics[width=1\columnwidth]{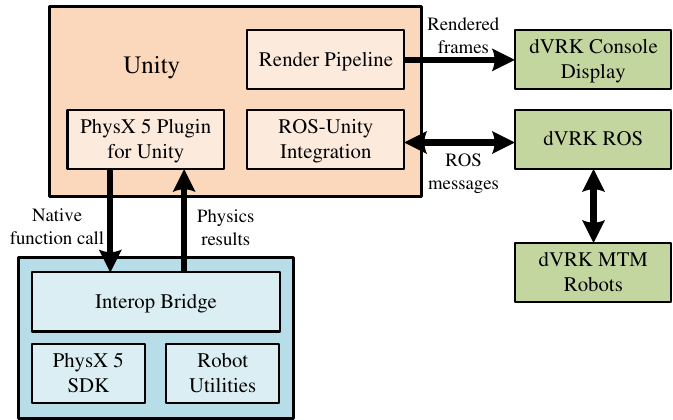}
    \caption{System architecture.}
    \label{fig:framework}%
\end{figure}

\section{System Architecture}
\subsection{Overview}
\label{sec:sys_overview}
The developed surgical simulator is based on Nvidia PhysX 5 SDK, the latest version of the Nvidia PhysX physics engine. Compared with previous versions that only support rigid body dynamics, PhysX 5 now supports a wide variety of physical simulations with GPU optimization, including (a) rigid bodies, (b) soft bodies using FEM, and (c) cloth, inflatables, and fluids using PBD. It also allows the simulation of serial robots using articulation joints. With PhysX 5 as the low-level physics engine, our simulation environment is built in Unity, a 3D game development software that has been widely used in surgical simulations. A Unity native plugin is developed to incorporate the PhysX engine. It is worth noting that the current built-in 3D physics engine of Unity is PhysX 4.1, which does not support FEM soft bodies and PBD-based objects. With the native support of PhysX 5 across platforms, we achieve cross-platform compatibility on both Windows and Linux.

\begin{figure}[htp]
    \centering
    \includegraphics[width=0.8\columnwidth]{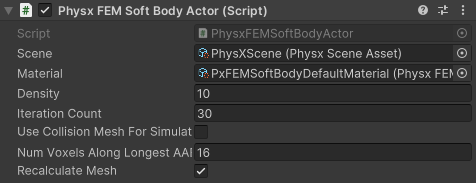}
    \caption{Inspector window for defining an FEM soft body.}
    \label{fig:inspector}%
\end{figure}

Fig.~\ref{fig:framework} shows an overview of the system architecture. \textit{PhysX 5 Plugin for Unity} is a Unity native plugin that and communicates with a native library \textit{Interop Bridge} to set the physics scene configuration and obtain the physics results. The plugin implements a number of custom script components for the Unity Inspector window to allow for GUI-based scene editing, as shown in Fig.~\ref{fig:inspector}. Unity's \textit{ROS-Unity Integration} package is used to achieve ROS communication with the \textit{dVRK-ROS} topics to interact with the physical master tool manipulator (MTM) robot from the dVRK. Using the \textit{Render Pipeline} provided by Unity, we can obtain stereo-vision frames and display them on the dVRK console displays.

\subsection{Physics Functionalities}
We utilize the physics functionalities provided by PhysX 5 to simulate a large variety of objects, including rigid bodies, serial robots, soft bodies, cloth, and fluid. To avoid undesirable behavior, Unity's built-in physics simulation which uses PhysX 4.1 is disabled.

\begin{figure}[ht]
    \centering
    \subfloat[]{\includegraphics[width=0.3\columnwidth]{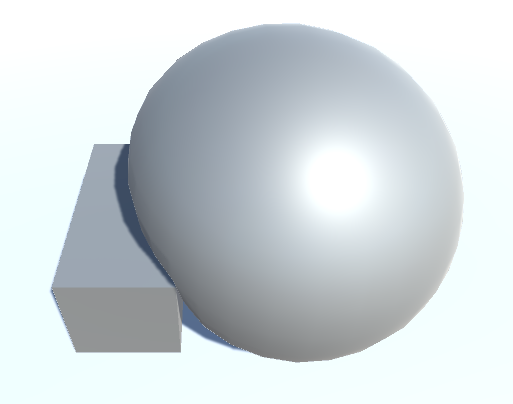}
    \label{fig:soft_rigid_demo}}
    \hfill
    \subfloat[]{\includegraphics[width=0.3\columnwidth]{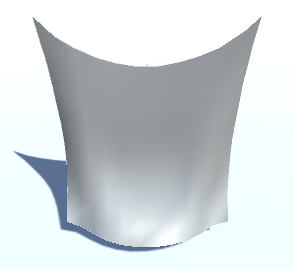}
    \label{fig:cloth_demo}}
    \hfill
    \subfloat[]{\includegraphics[width=0.3\columnwidth]{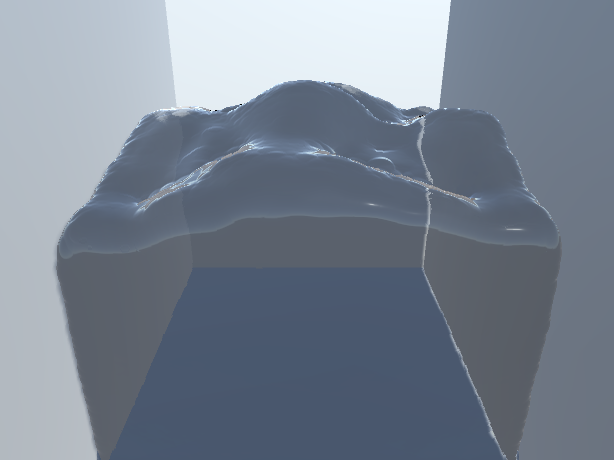}
    \label{fig:fluid_demo}}
    \hfill
    \caption{Examples of simulated objects. (a) Rigid and soft bodies; (b) Cloth with fixed vertices; (c) Fluid.}
    \label{fig:demos}%
\end{figure}

\subsubsection{Rigid body and robot}
\label{sec:rigid_body_robot}
PhysX provides a good foundation for rigid body dynamics. Taking advantage of this, CRESSim supports three types of rigid objects: static, dynamic, and kinematic rigid bodies. Furthermore, by utilizing the articulation joints, a type of joint that enables zero joint error using reduced coordinates, serial robots are supported by connecting rigid links through the articulation joints. Each articulation joint is driven by a PD controller. Common useful functions are implemented for serial robots, including forward kinematics, spatial and body Jacobians, and Jacobian-based numerical inverse kinematics. These are part of the \textit{Robot Utilities} component.

\subsubsection{FEM Soft body}
Leveraging GPU-based FEM soft body simulation provided by PhysX, we are able to simulate soft bodies such as tissue in surgical scenes. Parameters related to the physics properties of the soft body can be controlled, including Young's modulus, Poisson's ratio, and friction. Fig.~\ref{fig:soft_rigid_demo} shows a soft sphere colliding with a rigid box.

\subsubsection{PBD Cloth and Fluid}
The PBD particle system is also supported by PhysX with GPU. PBD is an efficient approach for simulating large amounts of particle system-based objects, including cloth and fluids, as shown in Fig.~\ref{fig:cloth_demo} and \ref{fig:fluid_demo}. In our implementation, both cloth and fluids include a set of 12 parameters for modifying the PBD material, such as friction, damping, and adhesion. In the case of fluid, buoyancy is an additional parameter that needs to be set. When using PBD cloth, each particle is connected with neighboring ones with constraints such as damped springs. Direct modification to the positions and velocities of the particles, as well as the spring constraints, are also supported during the simulation.

\subsubsection{Manipulation and Contact Features} Through direct access to the low-level PhysX library functions during simulation, a number of manipulation and customized contacts can be achieved. Using rigid-soft and rigid-particle attachments, we can achieve soft-body grasping and cloth grasping, which is common in surgical scenes. Furthermore, by removing the particle spring constraints in real time, cloth cutting can be efficiently simulated. Customized force fields that are applied to the particles are also implemented by modifying the particle positions and velocities at each simulation step.

\subsection{Teleoperation in Virtual Reality}

The real dVRK console display and the MTM robots are incorporated into the simulator to allow for teleoperating the simulated robots through VR. To allow communication between the simulator and the real MTM robots, \textit{ROS-Unity Integration} is used and the dVRK-specific ROS messages are generated in Unity. The process for single-arm teleoperation is summarized in Algorithm~\ref{alg:teleoperation}, where \texttt{AlignMTMWithPSM} is a Unity coroutine that waits until the MTM is positioned in the desired pose with the same orientation as the simulated patient side manipulator (PSM) robot. Each joint of the PSM robot is controlled by a PD controller. Since Unity's \texttt{FixedUpdate} function is used which is called every 0.02 seconds, the teleoperation is performed at a frequency of 50~Hz. During teleoperation, two cameras with a distance along the horizontal axis are present in the simulation scene to render two frames for stereo vision. The frames are shown on the two displays in the dVRK console, providing a VR experience. Fig.~\ref{fig:teleoperation_demo} shows the teleoperation of a simulated PSM in the blood suction scene by a human operator.

% ----- Algorithm 1 -----
\begin{algorithm}[hbt!]
\caption{Teleoperation using real MTMs}
\label{alg:teleoperation}
\begin{algorithmic}
\State \texttt{initializePSM $\gets$} true
\State \texttt{initializeMTM $\gets$} true
\State \texttt{isTeleoperating} $\gets$ false
\ForEach{\texttt{FixedUpdate} call}
    \If{\texttt{initializePSM}}
    \State Initialize PSM
    \State \texttt{initializePSM $\gets$} false
    \EndIf
    \If{\texttt{initializeMTM} and current time $\geq$ delay}
    \State Start \texttt{AlignMTMWithPSM} coroutine
    \State \texttt{initializeMTM $\gets$} false
    \If{\texttt{AlignMTMWithPSM} ends}
    \State \texttt{isTeleoperating} $\gets$ true
    \EndIf
    \EndIf
    \If{\texttt{isTeleoperating}}
    \State $\mathbf{P}_{PSM} \gets \mathbf{P}_{PSM} + scale \times \Delta\mathbf{P}_{MTM}$
    \State $\mathbf{R}_{PSM} \gets \mathbf{R}_{MTM}$
    \State $\mathbf{J} \gets$ \texttt{InverseKinematics($\mathbf{P}_{PSM}$, $\mathbf{R}_{PSM}$)}
    \State \texttt{DriveJoints($\mathbf{J}$)}
    \EndIf
\EndFor
\end{algorithmic}
\end{algorithm}

\section{Example Surgical Scenes}
\subsection{Simulated PSM with Various Surgical Instruments}
\begin{figure}[htp]
  \centering
  % Left sub-figure
  \subfloat[]{
    \includegraphics[width=0.8\columnwidth]{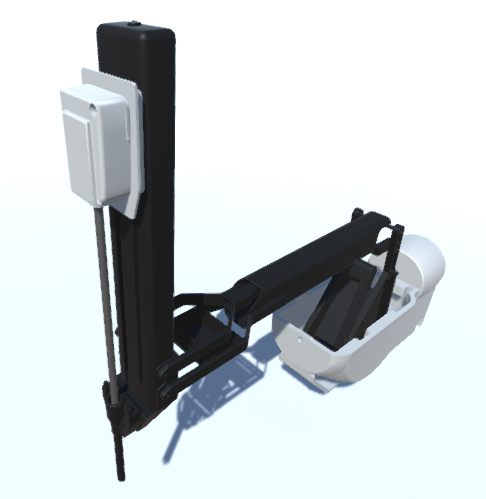}
    \label{fig:psm_demo}
  }
  \\
\subfloat[]{
    \includegraphics[height=0.25\columnwidth]{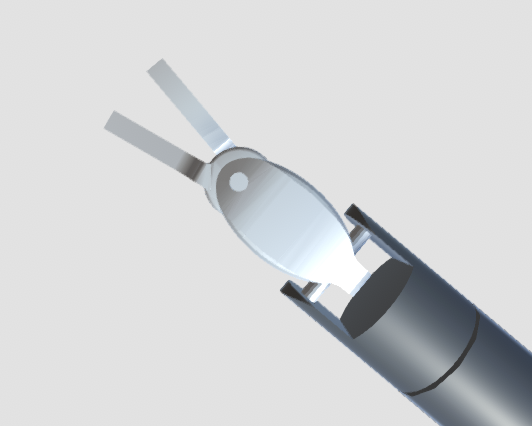}
    \label{fig:large_needle_driver_demo}
  }
\subfloat[]{
    \includegraphics[height=0.25\columnwidth]{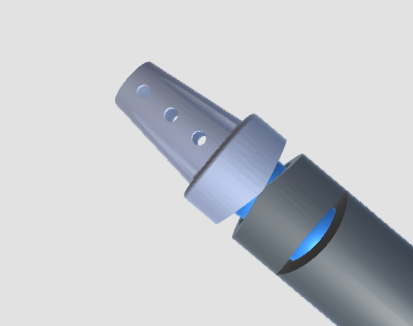}
    \label{fig:suction_irrigator_demo}
  }
\subfloat[]{
    \includegraphics[height=0.25\columnwidth]{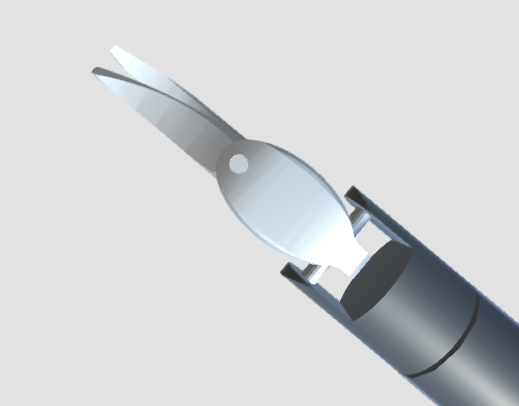}
    \label{fig:curved_scissor_demo}
  }
  \caption{Simulated PSM. (a) PSM robot; (b) Large needle driver; (c) Suction irrigator; (d) Curved scissor.}
\end{figure}

In this work, the PSM from the dVRK is simulated. The 3D models are modified from those in \cite{xu2021surrol} with improvements on mesh normals for rendering. As discussed in Section~\ref{sec:rigid_body_robot}, the PSM robot is defined by a series of rigid links through the reduced coordinate joints, with masses and inertia calculated from their bounding convex hull. Using the functionalities implemented in the \textit{Robot Utilities}, we are able to achieve both joint and Cartesian space control of the robot. The simulated PSM with the large needle driver is shown in Fig.~\ref{fig:psm_demo}.

Besides the large needle driver that has been simulated in various existing dVRK simulators, we additionally simulate two more surgical instruments: the curved scissor and the suction irrigator tool, as shown in Fig.~\ref{fig:large_needle_driver_demo} to \ref{fig:curved_scissor_demo}. These tools are used for simulating the complex contact-rich surgical tasks discussed in Section~\ref{sec:simulated_tasks}.

\begin{figure*}[ht]
    \centering
    \subfloat[]{\includegraphics[width=0.49\columnwidth]{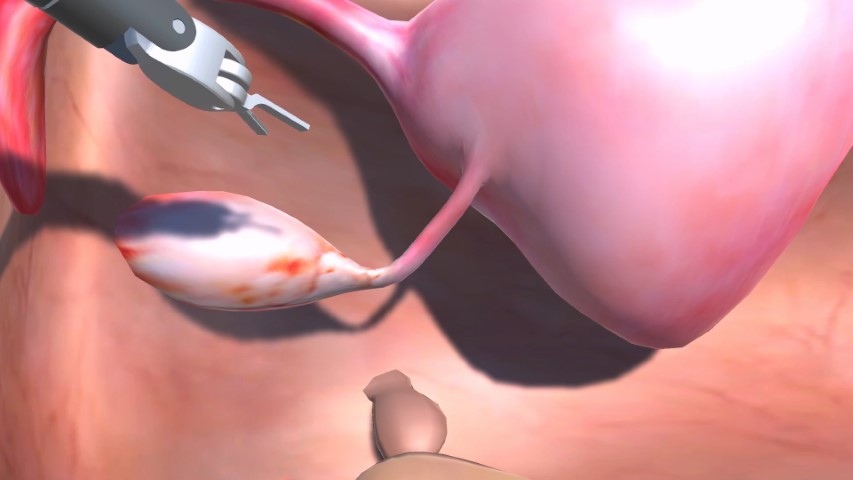}
    \label{fig:screenshots_grasping_main}}
    \hfill
    \subfloat[]{\includegraphics[width=0.49\columnwidth]{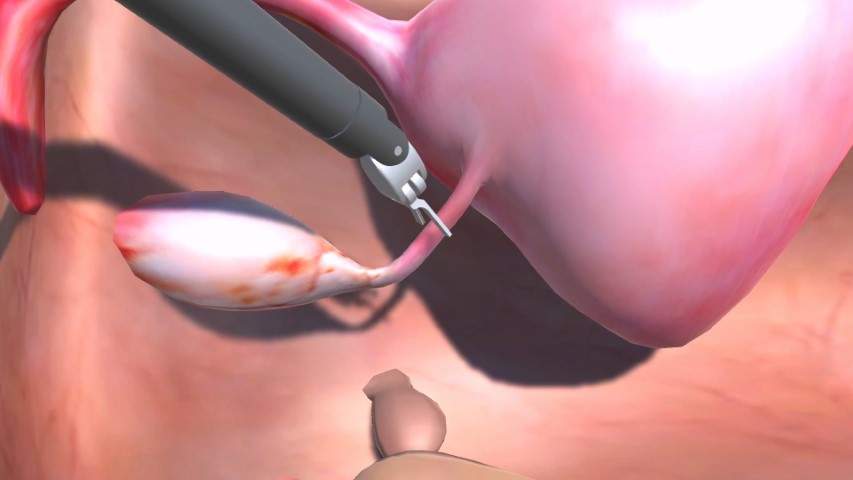}}
    \hfill
    \subfloat[]{\includegraphics[width=0.49\columnwidth]{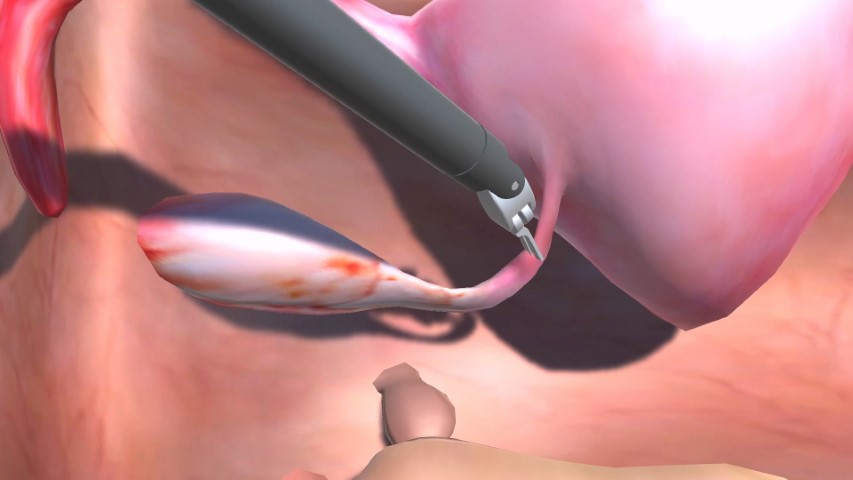}}
    \hfill
    \subfloat[]{\includegraphics[width=0.49\columnwidth]{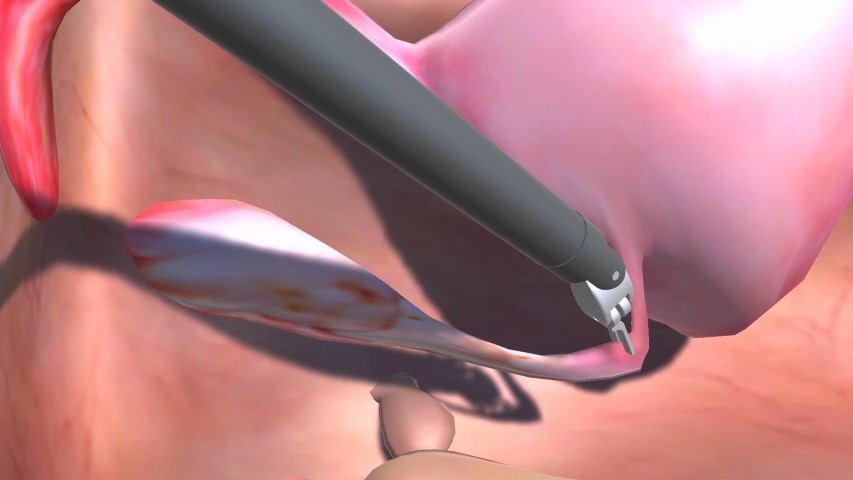}}    
    \caption{(a) to (d) is a sequence of screenshots taken from the tissue grasping and deformation task during teleoperation.}
    \label{fig:screenshots_grasping}%
\end{figure*}

\begin{figure*}[ht]
    \centering
    \subfloat[]{\includegraphics[width=0.49\columnwidth]{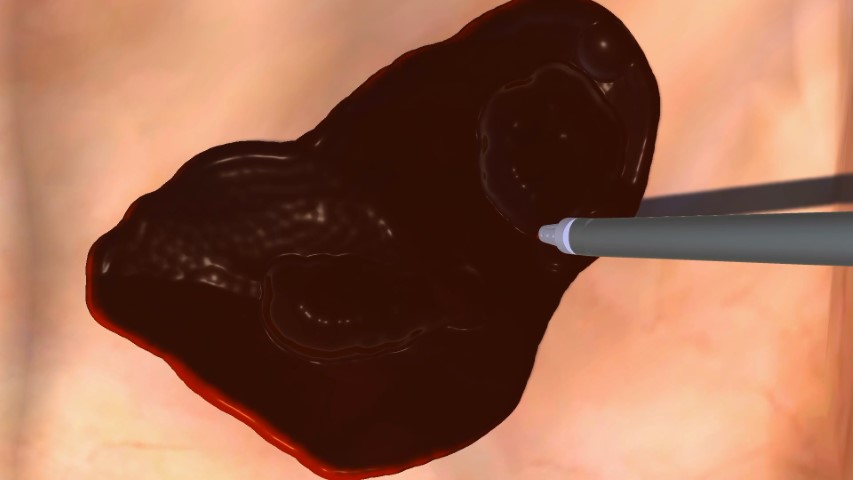}
    \label{fig:screenshots_blood_suction_main}}
    \hfill
    \subfloat[]{\includegraphics[width=0.49\columnwidth]{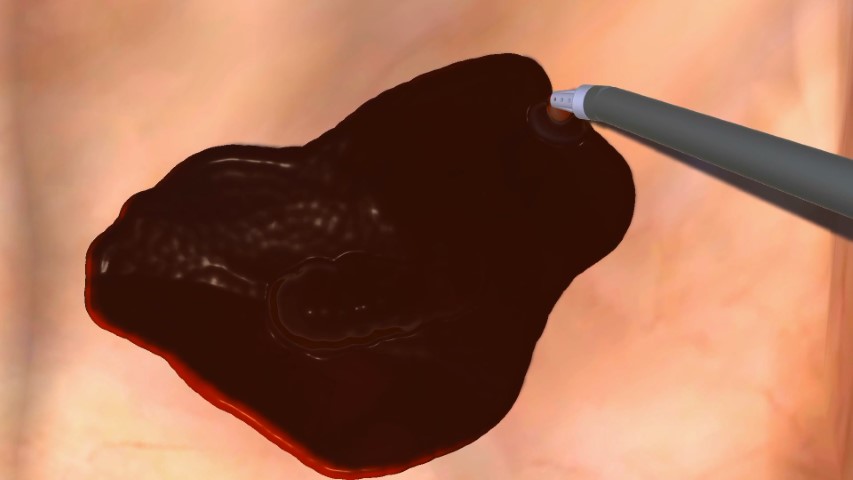}}
    \hfill
    \subfloat[]{\includegraphics[width=0.49\columnwidth]{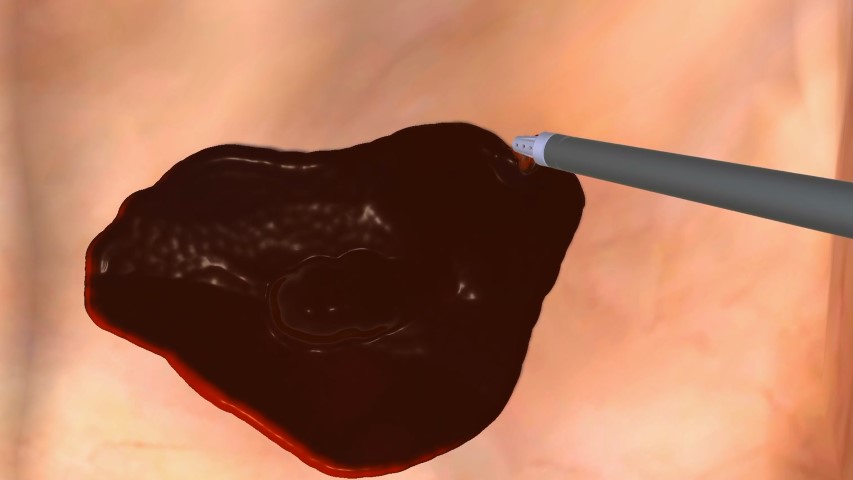}}
    \hfill
    \subfloat[]{\includegraphics[width=0.49\columnwidth]{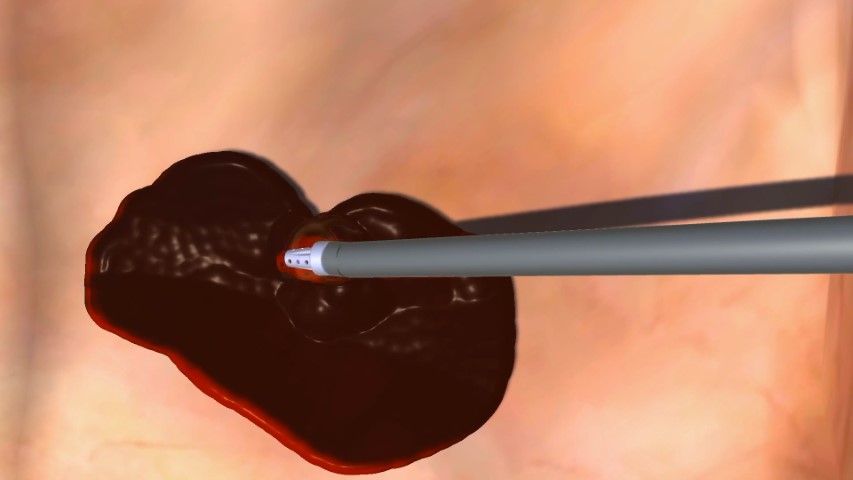}}   \\     
    \subfloat[]{\includegraphics[width=0.49\columnwidth]{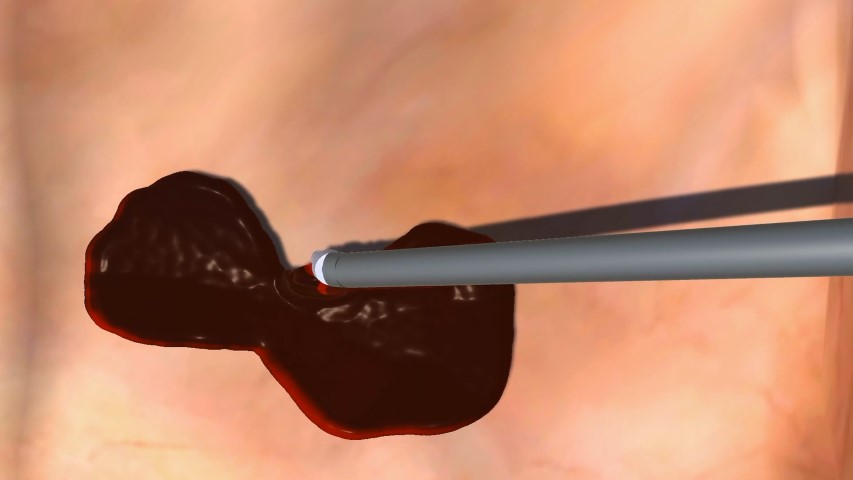}}
    \hfill
    \subfloat[]{\includegraphics[width=0.49\columnwidth]{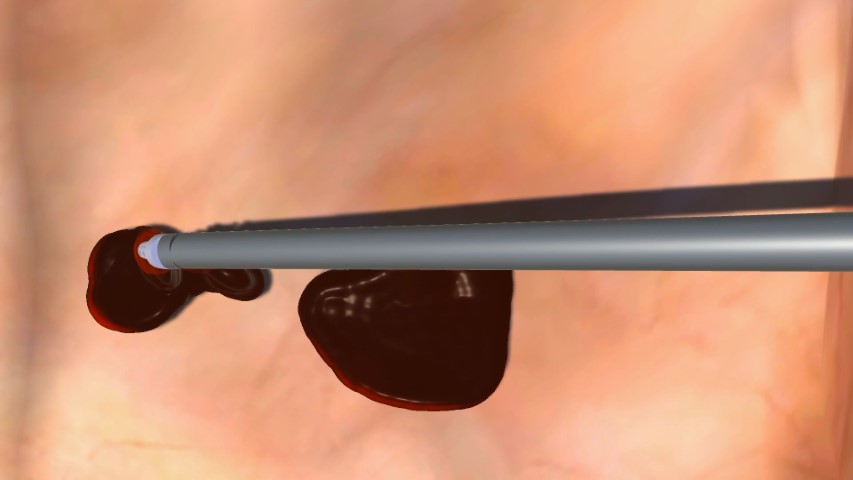}}
    \hfill
    \subfloat[]{\includegraphics[width=0.49\columnwidth]{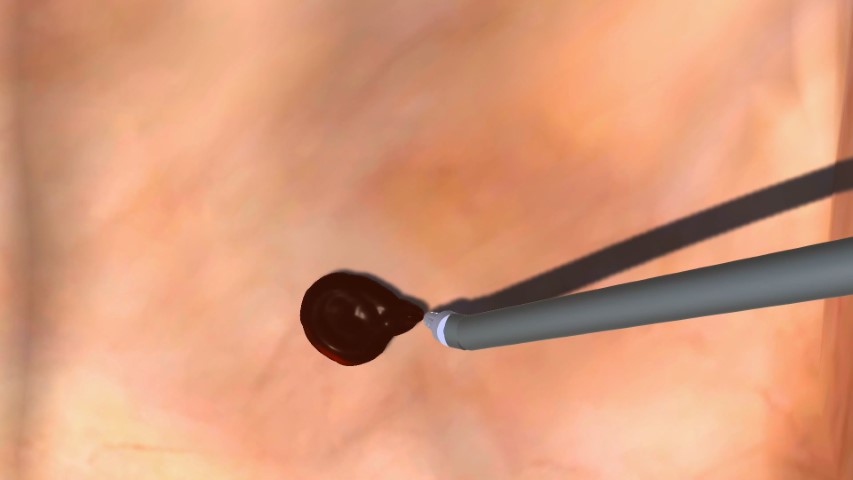}}
    \hfill
    \subfloat[]{\includegraphics[width=0.49\columnwidth]{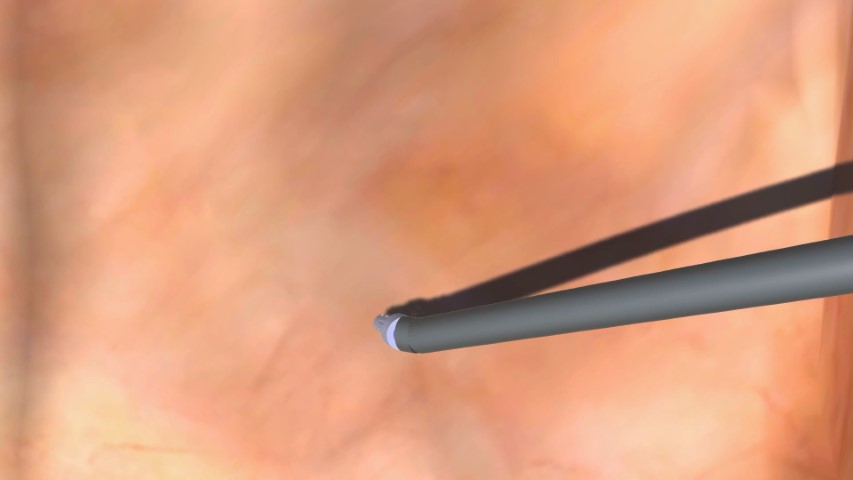}} 
    \caption{(a) to (h) is a sequence of screenshots taken from the blood suction task during teleoperation.}
    \label{fig:screenshots_blood_suction}%
\end{figure*}

\begin{figure*}[ht]
    \centering
    \subfloat[]{\includegraphics[width=0.49\columnwidth]{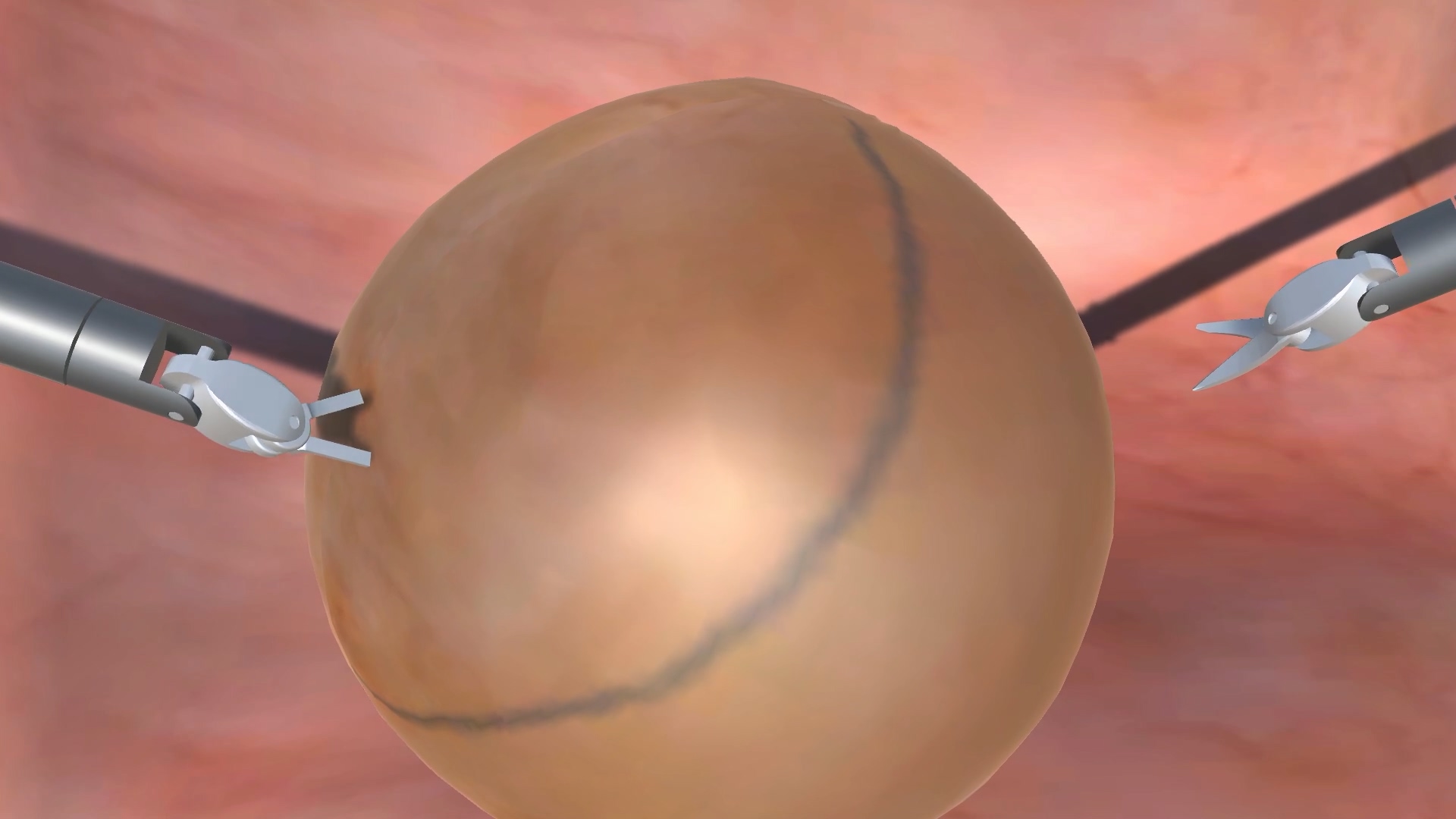}
    \label{fig:screenshots_cutting_main}}
    \hfill
    \subfloat[]{\includegraphics[width=0.49\columnwidth]{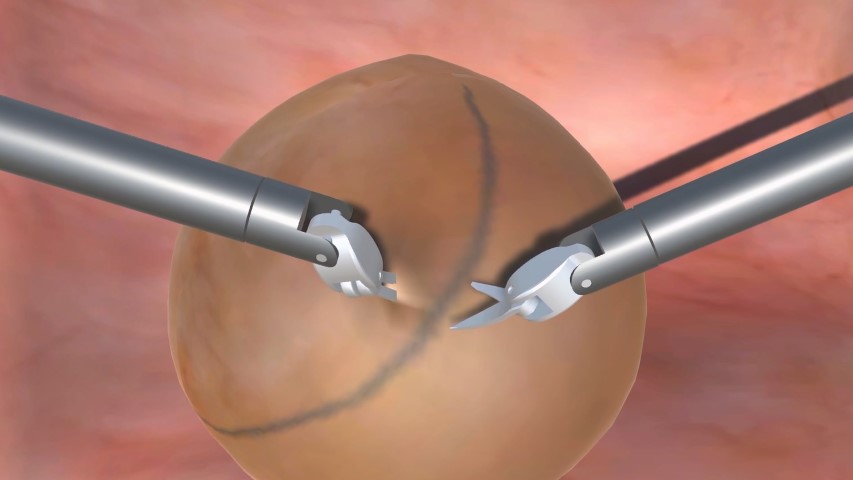}}
    \hfill
    \subfloat[]{\includegraphics[width=0.49\columnwidth]{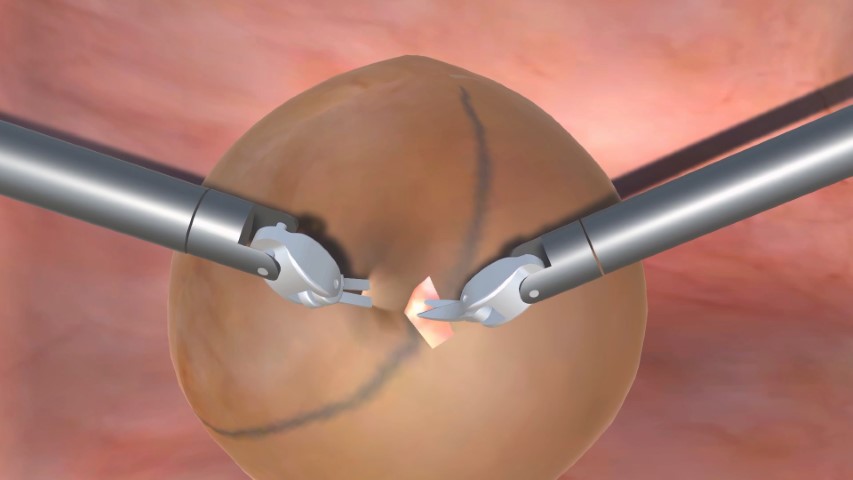}}
    \hfill
    \subfloat[]{\includegraphics[width=0.49\columnwidth]{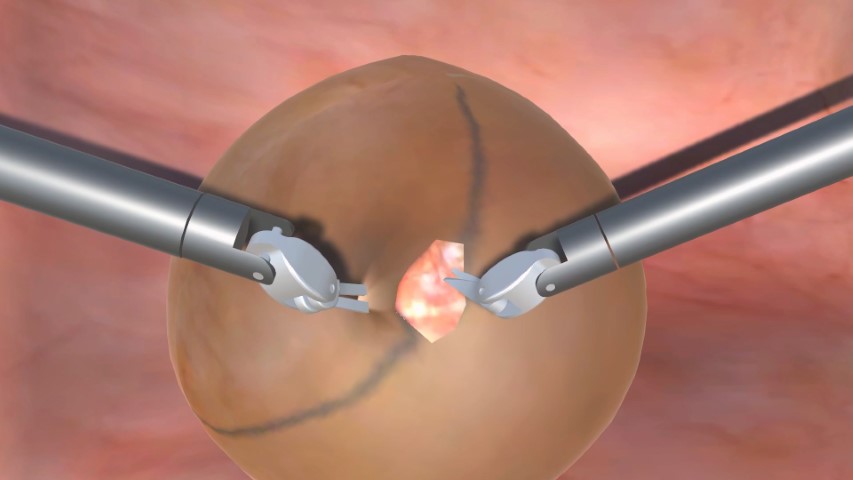}}   \\     
    \subfloat[]{\includegraphics[width=0.49\columnwidth]{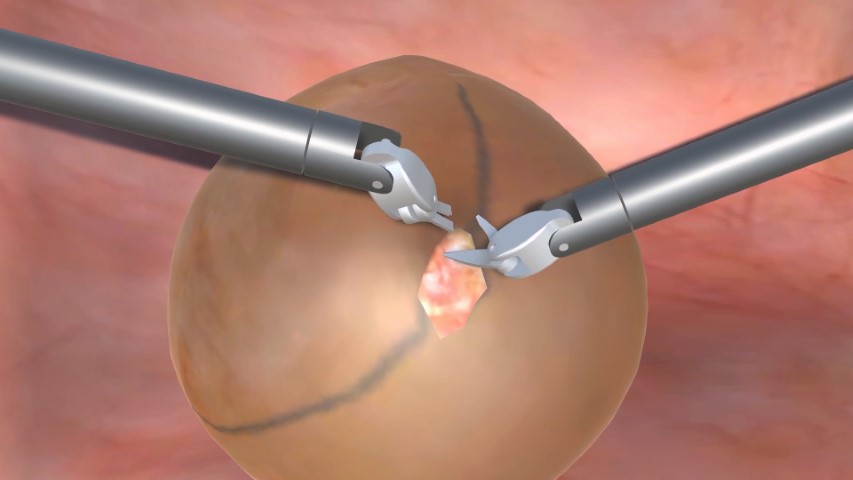}}
    \hfill
    \subfloat[]{\includegraphics[width=0.49\columnwidth]{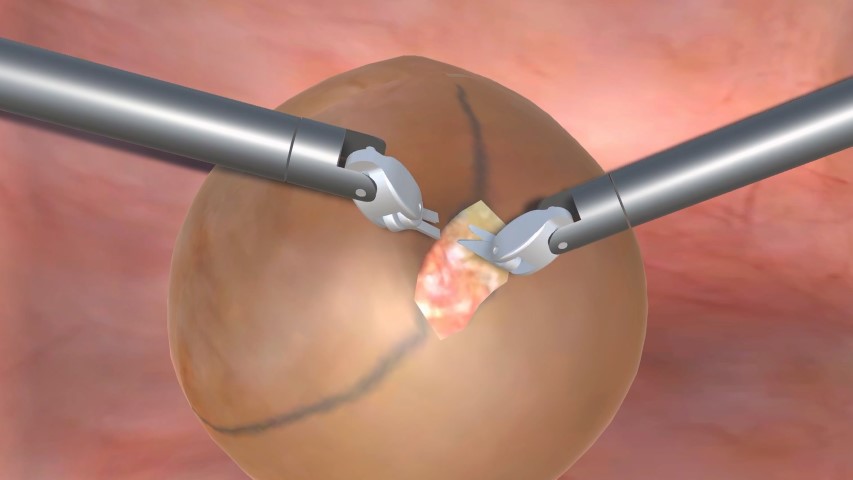}}
    \hfill
    \subfloat[]{\includegraphics[width=0.49\columnwidth]{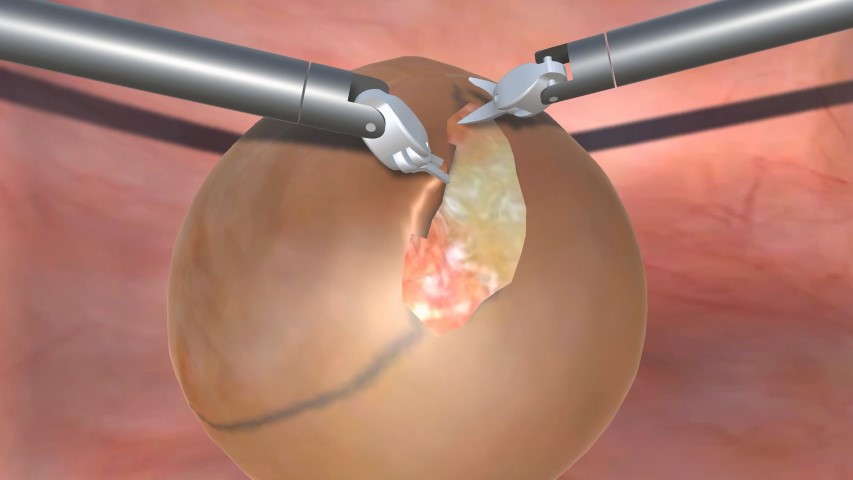}}
    \hfill
    \subfloat[]{\includegraphics[width=0.49\columnwidth]{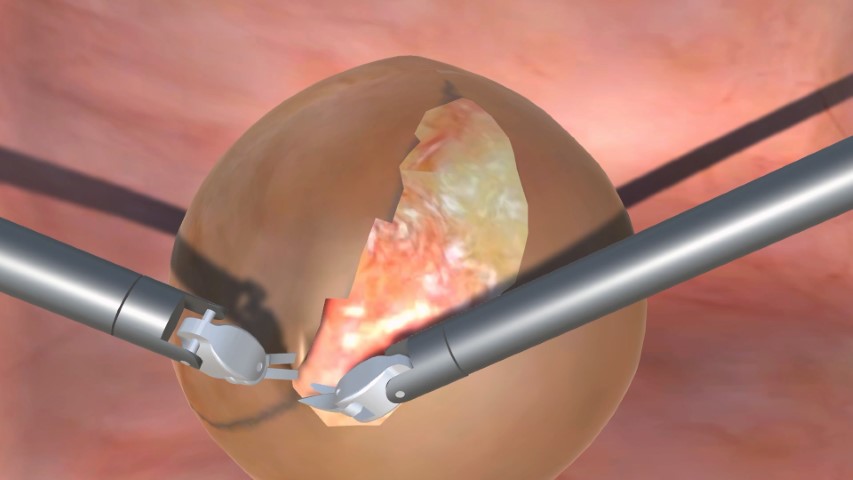}} 
    \caption{(a) to (h) is a sequence of screenshots taken from the tissue cutting task during teleoperation.}
    \label{fig:screenshots_cutting}%
\end{figure*}

% inverse and forward kinematics --> Cartesian control
\subsection{Simulated Contact-Rich Surgical Tasks}
\label{sec:simulated_tasks}
\subsubsection{Soft Tissue grasping and manipulation}
The task focuses on grasping and deforming soft tissue, a common sub-task in surgeries. We emulate the scene of a hysterectomy, where the uterus, ovary, and fat tissue are present, as shown in Fig.~\ref{fig:screenshots_grasping_main}. The goal of this task is to grasp and deform the organs to relocate some parts of the tissue for other sub-tasks such as cauterization, or to reveal the area underneath the tissue. The organs and tissue present in the scene are simulated as FEM soft bodies, and grasping is achieved by attaching a mesh vertex to the grasping tool. The task is inspired by \cite{tagliabue2020soft}, as well as a hysterectomy simulation scene in VirtaMed LaproS.

\subsubsection{Blood suction}
Blood suction is another common surgical sub-task where the goal is to use the suction irrigator tool to remove the blood in the scene, as shown in Fig.~\ref{fig:screenshots_blood_suction_main}. The background tissue is a large FEM soft body, and the blood is simulated by the PBD fluid. Suction is achieved by applying a customized force field around the suction tooltip. PBD particles are removed once they are close enough to the tooltip. This task is inspired by \cite{richter2021autonomous}.

\subsubsection{Tissue cutting}
In this task, a soft tissue with a thin layer of fascia is present and the goal is to grasp and pull the fascia and cut it along a given line, as shown in Fig.~\ref{fig:screenshots_cutting_main}. This task is bimanual and is commonly present in surgeries, although the cutting procedure may be replaced by cauterization. The background tissue is an FEM soft body. The spherical tissue for grasping and cutting consists of two parts: an FEM soft body inside which emulates the main part of the organ, and a thin layer of PBD cloth surrounding the organ, which simulates the fascia. The cloth layer has stretched spring constraints between particles to emulate the elastic behavior of a fascia. To simulate the cutting procedure, Algorithm~\ref{alg:tissue_cutting} is used to update the particle springs at each \texttt{FixedUpdate} function call, and the triangle mesh for rendering the tissue is updated accordingly. In practice, additional steps are needed to address issues related to triangle mesh welding. Furthermore, parallelism is needed for efficiently modifying the mesh in real time, due to the large number of particles and triangles needed for processing. The design of this task is inspired by similar tasks in da Vinci SimNow.

% ----- Algorithm 1 -----
\begin{algorithm}[hbt!]
\caption{Fascia tissue cutting}
\label{alg:tissue_cutting}
\begin{algorithmic}
\ForEach{\texttt{FixedUpdate} call}
    \ForEach{particle spring constraint \texttt{s}}
        \State $\mathbf{p}_0 \gets$ first particle position
        \State $\mathbf{p}_1 \gets$ second particle position
        \If{$\vv{\mathbf{p}_0\mathbf{p}_1}$ intersects with cutting curve}
        \State remove spring \texttt{s}
        \EndIf
    \EndFor
    \ForEach{render mesh triangles \texttt{t}}
    \If{no springs exist in at least two edges}
    \State remove triangle \texttt{t}
    \EndIf
    \EndFor
\EndFor
\end{algorithmic}
\end{algorithm}

To evaluate the quality of the designed simulation scenes, we perform the tasks using teleoperation and record the videos\footnote{\url{https://drive.google.com/drive/folders/1oHEhEt4K9kQYQMF5rYfPA6bthrXBM7gg}}. Please note that the scenes are designed to showcase the capability of the simulator in terms of the wide range of simulation objects, the achievable manipulation types, and the render capabilities. These scenes are not intended to be as accurate as possible to any specific surgical scenes, nor that they mimic the body of an actual patient, as designing actual surgical scenes requires extra 3D modeling of the tissue and organs, as well as art design of the materials and textures, which are out of the scope of this research.

Fig.~\ref{fig:screenshots_grasping} to \ref{fig:screenshots_cutting} show the screenshots taken from the teleoperation trials for three tasks. As shown in the figures, in all three tasks, we experienced close-to-real interactions in terms of the contact and collisions between different objects (the robots and the simulated tissue and organs) as well as the manipulation behavior. In soft tissue grasping and deformation, the tissue deforms in response to the motion of the robot grasper. In blood suction, blood is suctioned towards the suction tool and correctly removed from the scene. In tissue cutting, we are able to perform bimanual grasping and cutting on the tissue by grasping and stretching a point on the fascia using the left arm and cutting it along the line using the right arm. The contact and collisions between simulated objects significantly enhance the realism of the scenes, especially the collision between two robot arms, the collision between the robot arm and the soft tissue and organs, and the contact between soft tissue and organs.

\section{Discussion}
Although the physics computation for each \texttt{FixedUpdate} can take longer time when FEM and PBD objects are present, preliminary profiling results show that the simulator can run at least 50 to 60 frames per second (FPS), and we did not experience noticeable FPS drops during the teleoperation testing. Table~\ref{tab:profiling} shows the time spent on physics simulation at each step and the post-processing time for meshes and particles. Samples are taken from 10 consecutive physics updates, in which all \texttt{FixedUpdate} functions in scripts present in the scene are called. Tests are conducted on a PC with Intel Core i7-12700F and Nvidia RTX 3070. As expected, blood suction and tissue cutting scenes add more physics computation time due to the added complexity of the scene with both FEM and PBD objects, and the additional time needed for processing the meshes and particles. However, as the code is not fully optimized for computational efficiency, there is room for improvement in future work, for example, by adding more parallelism.
We have also noticed slight delays in teleoperation, primarily because the PD gains for the robot joint controllers are not well-tuned to rapidly drive the joint angles to the setpoint. This will be addressed in future work as well.

\begin{table*}[ht]
\centering
\caption{Physics simulation and post-processing time.}
\label{tab:profiling}
\begin{tabular}{cccc}
\hline
Scenes & Physics advance (ms) & Mesh and particle post-processing (ms) & \texttt{FixedUpdate} total (ms) \Tstrut\Bstrut \\ \hline
Tissue grasping and deformation & $9.51\pm0.63$ & $0.12\pm0.02$ & $9.68\pm0.63$ \Tstrut\\
Blood suction & $14.00\pm0.31$ & $0.19\pm0.03$ & $14.24\pm0.31$ \\
Tissue cutting & $15.86\pm0.39$ & $0.39\pm0.16$ & $16.52\pm0.81$ \Bstrut\\ \hline
\end{tabular}
\end{table*}

With Unity as an intuitive platform for defining the simulated scenes through GUI, it is straightforward to further create realistic scenes for real surgical tasks, as long as the tissue and surgical instruments are represented in 3D models.
The long-term objective of CRESSim is to emulate the functionalities provided by the da Vinci SimNow and VirtaMed LaproS to allow for simulating various realistic surgical tasks, but with a particular focus on surgical robotics research instead of commercial usage. The source code will be made available to the community to facilitate research in surgical automation and autonomy, as well as other applications that require realistic surgical task simulations. One specific application of the simulator is simulation-to-reality surgical robot learning using reinforcement learning and imitation learning algorithms.

While this work presents an initial implementation and three simulated scenes, we have not been able to conduct user studies on the quality and realism of the simulator. Future studies could include evaluations and feedback from surgeons, especially from users of the da Vinci Surgical System and the da Vinci SimNow. Additionally, other limitations need to be addressed in the future development of CRESSim. One major limitation is that there are a large number of surgical instruments and procedures, such as cauterization, that have not been covered in this work. Furthermore, although cutting can be simulated for the PBD cloth, which can be used to simulate cutting thin tissue such as fascia, FEM soft body cutting is not achieved. Simulating cutting for the FEM soft body requires extensive real-time re-calculation of the tetrahedron mesh, making it extremely challenging. This limits the simulation of cutting thick and voluminous soft tissue in surgeries, which is also a common procedure. We expect to address these limitations in the future work. 

\section{Conclusion}
In this work, we present CRESSim, a realistic surgical simulation environment for the dVRK that allows contact-rich manipulations. The simulator is built on Unity and PhysX 5 SDK, and the real dVRK console and MTMs are incorporated into the simulator for VR-based teleoperation. Thanks to the simulation capabilities provided by PhysX~5 compared with its previous versions, we are able to simulate soft bodies, cloth, inflatables, and fluids that exist in surgeries, and the contact-rich manipulation of these objects. To show the advantages and potentials of the developed simulator, we showcase 3 examples of surgical simulation scenes and their corresponding tasks, including tissue grasping and deformation, blood suction, and tissue cutting. Preliminary experiments and profiling show the platform's capability to simulate surgical tasks and allow real-time teleoperation. In future work, we will further enhance the simulator to cover more realistic surgical scenes and various surgical instruments.

\addtolength{\textheight}{-12cm}   % This command serves to balance the column lengths


% \section*{ACKNOWLEDGMENT}

\bibliographystyle{IEEEtran}
\bibliography{references.bib}

\end{document}